\documentclass[runningheads]{llncs}
\usepackage{graphicx}

\usepackage{tikz}
\usepackage{multirow}
\usepackage{algorithm}
\usepackage{algpseudocode}
\usepackage{bbm}
\usepackage{comment}
\usepackage{amsmath,amssymb} 
\usepackage{color}

\usepackage[accsupp]{axessibility}  

\begin{document}
\pagestyle{headings}
\mainmatter
\def\ECCVSubNumber{4}  

\title{Reliable Multimodal Trajectory Prediction via Error Aligned Uncertainty Optimization} 

\titlerunning{Error Aligned Uncertainty Optimization}
%

\author{Neslihan Kose \and
Ranganath Krishnan \and
Akash Dhamasia \and
Omesh Tickoo \and
Michael Paulitsch}
\authorrunning{N. Kose et al.}
%
\institute{Intel Labs \\
\email{\{neslihan.kose.cihangir, ranganath.krishnan, akash.dhamasia, omesh.tickoo, michael.paulitsch\}@intel.com}} 
\maketitle

\begin{abstract}
Reliable uncertainty quantification in deep neural networks is very crucial in safety-critical applications such as automated driving for trustworthy and informed decision-making. Assessing the quality of uncertainty estimates is challenging as ground truth for uncertainty estimates is not available. Ideally, in a well-calibrated model, uncertainty estimates should perfectly correlate with model error. We propose a novel error aligned uncertainty optimization method and introduce a trainable loss function to guide the models to yield good quality uncertainty estimates aligning with the model error. Our approach targets continuous structured prediction and regression tasks, and is evaluated on multiple datasets including a large-scale vehicle motion prediction task involving real-world distributional shifts. We demonstrate that our method improves average displacement error by $1.69\%$ and $4.69\%$, and the uncertainty correlation with model error by $17.22\%$ and $19.13\%$ as quantified by Pearson correlation coefficient on two state-of-the-art baselines.

\end{abstract}

\section{Introduction}\label{sec:intro}

Conventional deep learning models often tend to make unreliable predictions~\cite{guo2017calibration,Ovadia2019} and do not provide a measure of uncertainty in regression tasks. Incorporating uncertainty estimation to deep neural networks enables informed decision-making as it indicates how certain the model is with its' prediction. In safety-critical applications such as automated driving (AD), in addition to making accurate predictions, it is important to quantify uncertainty associated with predictions~\cite{ghahramani2015probabilistic} in order to establish trust in the models. 

A well-calibrated model should yield low uncertainty when the prediction is accurate, and indicate high uncertainty for inaccurate predictions. 
Since the ground truths for uncertainty estimates are not available, uncertainty calibration is challenging, and has been explored mostly for classification tasks or post-hoc fine-tuning so far. In this paper, we propose a novel \textbf{error aligned uncertainty optimization} technique introducing a differentiable secondary loss function to obtain well-calibrated models for regression-based tasks including the challenging continuous structured prediction task such as vehicle motion prediction.

\begin{figure*}[t!]
  \centering
  \includegraphics[width=1\linewidth,page=3]{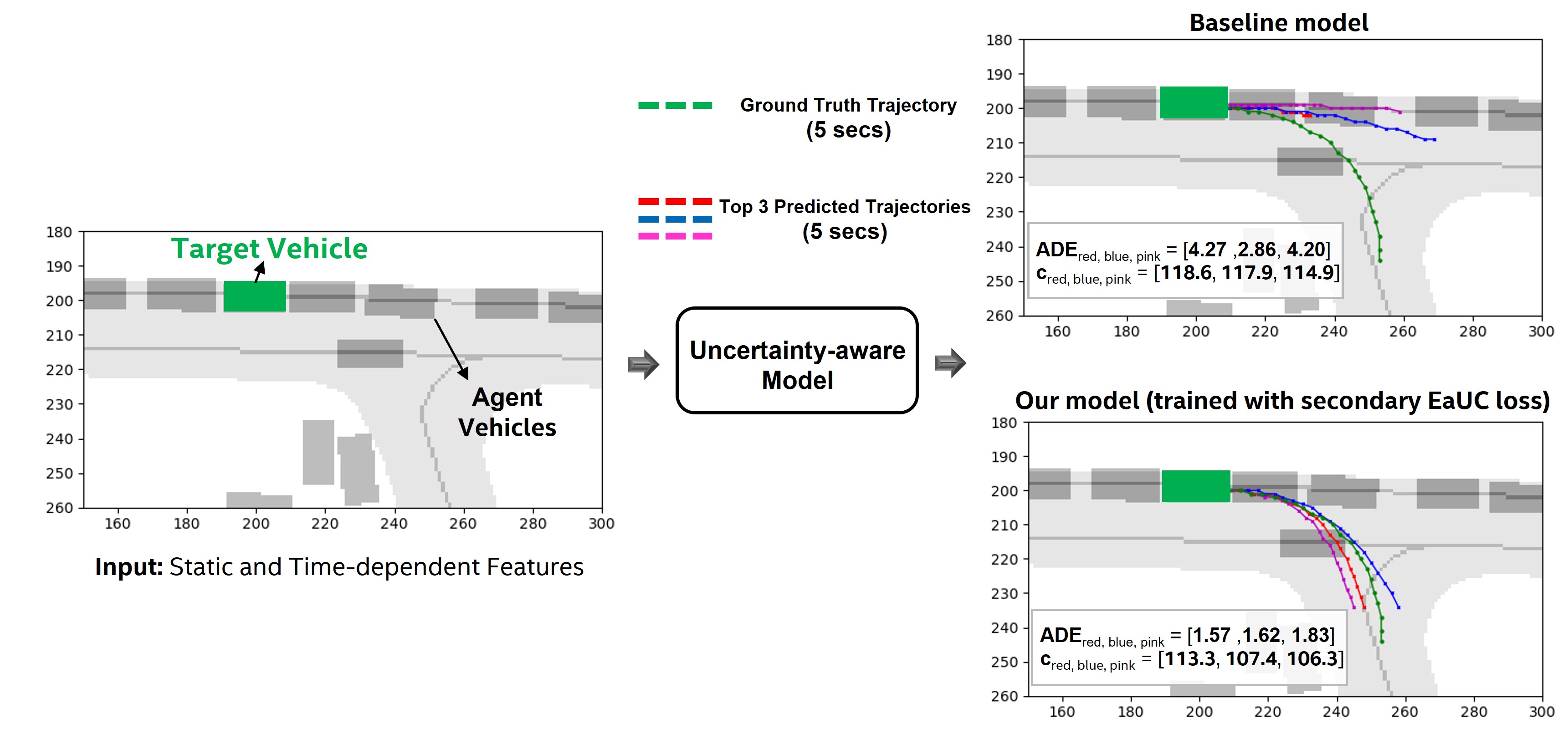}
  \caption{Figure shows how our approach improves model calibration when it is incorporated to state-of-the-art uncertainty-aware vehicle trajectory prediction baseline (Deep Imitative Model~\cite{DIM18}) on Shift Dataset~\cite{shifts2021}. Here, $c$ and $ADE$ denote log-likelihood score (certainty measure) and average displacement error (robustness measure), respectively. In well-calibrated models, we expect higher $c$ value for samples with lower $ADE$ and lower $c$ value for samples with higher $ADE$.}\label{fig:flowchart}
\end{figure*}

Fig. \ref{fig:flowchart} shows how our approach improves model calibration when it is incorporated to state-of-the art (SoTA) uncertainty-aware vehicle trajectory prediction model, which involves predicting the next trajectories (i.e., possible future states) of agents around the automated vehicle. Vehicle trajectory prediction is a crucial component of AD. This area has recently received considerable attention from both industry and academia~\cite{huang2019uncertainty,VectorNet20,ECCV20,CVPR21}. Reliable uncertainty quantification for prediction module is of critical importance for the planning module of autonomous system to make the right decision leading to safer decision-making. Related safety standards like ISO/PAS 21448~\cite{iso2019} suggest uncertainty is an important safety aspect. In addition to being a safety-critical application, trajectory prediction is also a challenging task for the uncertainty estimation domain due to the structured and continuous nature of predictions. 

In this paper, our contributions are as follows: \begin{itemize}
    \item We propose a novel error aligned uncertainty (EaU) optimization method and introduce EaU calibration loss to guide the models to provide reliable uncertainty estimates correlating with the model error. To the best of our knowledge, this is the first work to introduce a trainable loss function to obtain reliable uncertainty estimates  for continuous structured prediction (e.g., trajectory prediction) and regression tasks. 
    \item We evaluate the proposed method with the large-scale real-world Shifts dataset and benchmark~\cite{shifts2021} for vehicle trajectory prediction task using two different baselines. We also use UCI datasets~\cite{Dua2019} as an additional evaluation of our approach considering multiple regression tasks. The results from extensive experiments demonstrate our method improves SoTA model performance on these tasks and yields well-calibrated models in addition to improved robustness even under real-world distributional shifts. 
\end{itemize}

In the rest of the paper, our approach is explained considering vehicle motion prediction as it is a safety-critical application of AD, and challenging application of uncertainty quantification due to the continuous, structured and multimodal nature of predictions. Here, multimodality refers to multiple trajectory predictions that are plausible for a target vehicle in a certain context. Our method is applicable to any regression task, which we later show on UCI benchmark.

\section{Related Work}

This paper focuses on reliable uncertainty quantification for continuous structured prediction tasks such as trajectory prediction in addition to any regression-based tasks introducing a novel differentiable loss function that correlates uncertainty estimates with model error.

There are various approaches to estimate uncertainty in neural network predictions including Bayesian~\cite{welling2011bayesian,blundell2015weight,gal2016dropout,zhang2020csgmcmc,pmlr2020BNN} and non-Bayesian~\cite{lakshminarayanan2017simple,liu2020simple,pmlr2020} methods. Our proposed error aligned uncertainty optimization is orthogonal to and can complement existing uncertainty estimation methods as it can be applied as a secondary loss function together with any of these methods to get calibrated good quality uncertainty estimates in addition to improved robustness.

The existing solutions to get well-calibrated uncertainties are mainly developed for classification tasks or based on post-hoc calibration so far.

In \cite{AvUC20}, the authors propose an optimization method that is based on the relationship between accuracy and uncertainty. For this purpose, they introduced differentiable accuracy versus uncertainty calibration (AvUC) loss which enables the model to provide well-calibrated uncertainties in addition to improved accuracy. The paper shows promising results, however it is applicable only for classification tasks. We follow the insights from AvUC to build a loss-calibrated inference method for time-series based prediction and regression tasks.

Post-hoc methods~\cite{Kuleshov18,kumar2019verified,UAI21} are based on recalibrating a pretrained model with a sample of independent and identically distributed (i.i.d.) data, and have been devised mainly for classification or small-scale regression problems. \cite{Kuleshov18} proposed a post-hoc method to calibrate the output of any 1D regression algorithm based on Platt scaling~\cite{Platt99probabilisticoutputs}. In \cite{UAI21}, a truth discovery framework is proposed integrating ensemble-based and post-hoc calibration methods based on an accuracy preserving truth estimator. The authors show that post-hoc methods can be enhanced by truth discovery-regularized optimization. To the best of our knowledge, there is no existing post-hoc calibration solution for the time-series based multi-variate regression tasks such as trajectory prediction as they bring additional challenges when designing these methods. \cite{Ovadia2019} has also shown that post-hoc methods fail to provide well-calibrated uncertainties under distributional shifts in the real-world.

Recently, Shifts dataset and benchmark \cite{shifts2021} were introduced for uncertainty and robustness to distributional shift analysis on multimodal vehicle trajectory prediction. Multimodal techniques in this context output a distribution over multiple trajectories ~\cite{Chai2019,Cui2019,Minh2020,Hong2019,Makansi2019}. The Shifts dataset is currently the largest publicly available vehicle motion prediction dataset with 600,000 scenes including real-world distributional shifts as well. The Shifts benchmark incorporates uncertainties to vehicle motion prediction pipeline and proposes metrics and solutions on how to jointly assess robustness and uncertainties in this task. In \cite{shifts2021}, the authors emphasize that in continuous structured prediction tasks such as vehicle trajectory prediction, there is still much potential for further development of informative measures of uncertainty and well-calibrated models.

To the best of our knowledge, this is the first work proposing an orthogonal solution based on a new trainable secondary loss function that can complement existing uncertainty estimation methods for continuous structured prediction tasks to improve the quality of their uncertainty measures.

\section{Setup for Vehicle Trajectory Prediction}

Vehicle trajectory prediction aims to predict the future states of agents in the scene. Here, the training set is denoted as $\mathcal{D}_{train}={(x_{i},y_{i})}_{i=1}^{N}$. $y$ and $x$ denote the ground truth trajectories and high-dimensional features representing scene context, respectively. Each $y = (s_{1}, ..., s_{T})$ denotes the future trajectory of a given vehicle observed by the automated vehicle perception stack. Each state $s_{t}$ corresponds to the $d_{x}$- and $d_{y}$-displacement of the vehicle at timestep $t$, where $y \in R^{T\times2}$ (Fig. \ref{fig:flowchart}). 

Each scene has $M$ seconds duration. It is divided into $K$ seconds of context features and $L$ seconds of ground truth targets for prediction that is separated by the time $T = 0$. The goal is to predict the future trajectory of vehicles at time $T \in (0, L]$ based on the information available for time $T \in [-K, 0]$.

In \cite{shifts2021}, the authors propose a solution on how to incorporate uncertainty to motion prediction, and introduce two types of uncertainty quantification metrics: \begin{itemize}
    \item \textit{Per-trajectory confidence-aware metric:} For a given input $x$, the stochastic model accompanies its $D$ top trajectory predictions with scalar per-trajectory confidence scores $(c^{(i)} |i \in 1,.., D)$. 
    \item \textit{Per-prediction request uncertainty metric:} $U$ is computed by aggregating the $D$ top per-trajectory confidence scores to a single uncertainty score, which represents model’s uncertainty in making any prediction for the target agent.
\end{itemize}
In our paper, per-trajectory metric $c$ represents our certainty measure based on the log-likehood (Fig. \ref{fig:flowchart}). Per-prediction request uncertainty metric $U$ is computed with negating the mean average of $D$ top per-trajectory certainty scores.

As shown in Fig. \ref{fig:flowchart}, the input of the model is a single scene context $x$ consisting of static (map of the environment that can be augmented with extra information such as crosswalk occupancy, lane availability, direction, and speed limit) and time-dependent input features (e.g., occupancy, velocity, acceleration and yaw for vehicles and pedestrians in the scene). The output of the model is $D$ top trajectory predictions ($y^{(d)}|d \in 1,.., D$) for the future movements of the target vehicle together with their corresponding certainty ($c^{(d)} |d \in 1,.., D$) as in Fig. \ref{fig:flowchart} or uncertainty ($u^{(d)} |d \in 1,.., D$) scores as well as a single per-prediction request uncertainty score $U$. In the paper, we use $c$ and $u$ as our uncertainty measure interchangeably according to the context. Higher $c$ indicates lower $u$.

In the existence of unfamiliar or high-risk scene context, an automated vehicle associates a high per-prediction request uncertainty \cite{shifts2021}. However, since uncertainty estimates do not have ground truth, it is challenging to assess their quality. In our paper, we jointly assess the quality of uncertainty measures and robustness to distributional shift according to the following problem dimensions: 

\textit{Robustness to distributional shift} is mainly assessed via metrics such as Average Displacement Error (ADE) or Mean Square Error (MSE) in case of continuous structured prediction and regression tasks, respectively. ADE is the standard performance metric for time-series data and measures the quality of a prediction $y$ with respect to the ground truth $y^{*}$ as given in Eq. \ref{eq:ADE}, where $y = (s_{1}, ... , s_{T})$.\begin{equation}\label{eq:ADE}
 ADE(\textbf{y}) := \frac{1}{T} \sum_{t=1}^{T} ||s_{t}-s_{t}^{*}||_{2}.
\end{equation}
The analysis is done with two types of evaluation datasets, which are the in-distribution and shifted datasets. Models are considered more robust in case smaller degradation in performance is observed on the shifted data.

\textit{Quality of uncertainty estimates} should be jointly assessed with robustness measure as there could be cases such that the model performs well on shifted data (e.g, for certain examples of distributional shift) and poorly on in-distribution data (e.g., on data from a not well-represented part of the training set) \cite{shifts2021}. Joint assessment enables to understand whether measures of uncertainty correlate well with the presence of an incorrect prediction or high error. 

Our goal is to have reliable uncertainty estimates which correlate well with the corresponding error measure. For this purpose, we propose an error aligned uncertainty (EaU) optimization technique to get well-calibrated models in addition to improved robustness.

\section{Error Aligned Uncertainty Optimization}

In order to develop well-calibrated models for continuous structured prediction and regression tasks, we propose a novel \textbf{Error aligned Uncertainty Calibration (EaUC)} loss, which can be used as a secondary loss utility function relying on theoretically sound Bayesian decision theory~\cite{bayesian85}. A task specific utility function is used for optimal model learning in Bayesian decision theory. EaUC loss is intended to be used as a secondary task specific utility function to achieve the objective of training the model to provide well-calibrated uncertainty estimates. 

We build our method with the insights from AvUC loss\cite{AvUC20} that was introduced for classification tasks, and extend previous work on classification to continuous structured prediction and regression tasks.

In regression and continuous structured prediction tasks, robustness is measured typically in terms of Mean Square Error (MSE) and Average Displacement Error (ADE), respectively, instead of accuracy score. Therefore our robustness measure is ADE here as we explain our technique on motion prediction task. Lower MSE and ADE indicate more accurate results. 

\begin{minipage}{0.4\textwidth}
    \begin{flushleft}
    \begin{tabular}{cccc} 
    \cline{3-4}
    & & \multicolumn{2}{ c }{Uncertainty} \\ 
    & & \multicolumn{1}{ c|} {certain} & uncertain \\ \cline{1-4}
    \multicolumn{1}{ c  }{\multirow{2}{*}{ADE}} & low & \multicolumn{1}{ |c|}{LC} & LU \\ \cline{2-4}
    \multicolumn{1}{ c  }{} & high & \multicolumn{1}{ |c|}{HC} & HU \\ \cline{1-4} 
    \end{tabular}
    \end{flushleft}
\end{minipage}
\begin{minipage}{0.4\textwidth}
    \begin{equation*}
      EaU = \frac{n_{LC} + n_{HU}}{n_{LC} + n_{LU} + n_{HC} + n_{HU}}.
    \end{equation*}
\end{minipage}
\begin{minipage}{0.15\textwidth}
    \begin{equation} \label{eq:EAU}
    \end{equation}
\end{minipage}

In the rest of this paper, we use the notations in Eq. \ref{eq:EAU} to represent the number of accurate \& certain samples ($n_{LC}$), inaccurate \& certain samples ($n_{HC}$), accurate \& uncertain samples ($n_{LU}$) and inaccurate \& uncertain samples ($n_{HU}$). Here, $L$ and $H$ refer to low and high error, respectively, and $C$ and $U$ refer to certain and uncertain sample, respectively.

Ideally, we expect the model to be certain about its prediction when it is accurate and provide high uncertainty when making inaccurate predictions (Fig. \ref{fig:flowchart}). So, our aim is to have more certain samples when the predictions are accurate ($LC$), and uncertain samples when they are inaccurate ($HU$) compared to having uncertain samples when they are accurate ($LU$), and certain samples when they are inaccurate ($HC$). For this purpose, we propose the Error aligned Uncertainty (EaU) measure shown in Eq. \ref{eq:EAU}. A reliable and well-calibrated model provides higher EaU measure ($EAU \in [0, 1]$).

Eq. \ref{eq:EaU1} shows how we assign each sample to the corresponding class of samples. 
\begin{equation}
\label{eq:EaU1}
\begin{split}
  n_{LU} := \sum_{i} \mathbbm{1} (ade_{i} \leq ade_{th} \quad \textrm{and} \quad c_{i} \leq {c_{th}}); \\
  n_{HC} := \sum_{i} \mathbbm{1} (ade_{i} > ade_{th} \quad \textrm{and} \quad c_{i} > {c_{th}}); \\
  n_{LC} := \sum_{i} \mathbbm{1} (ade_{i} \leq ade_{th} \quad \textrm{and} \quad c_{i} > {c_{th}}); \\
  n_{HU} := \sum_{i} \mathbbm{1} (ade_{i} > ade_{th} \quad \textrm{and} \quad c_{i} \leq {c_{th}}); \\
\end{split}
\end{equation}

In Eq. \ref{eq:EaU1}, we use average displacement error, $ade_{i}$, as our robustness measure to classify the sample as accurate or inaccurate comparing it with a task-dependant threshold $ade_{th}$. The samples are classified as certain or uncertain according to their certainty score $c_{i}$ that is based on log-likelihood in our motion prediction use case. Similarly, log-likelihood of each sample, which is our certainty measure, is compared with a task-dependant threshold $c_{th}$.  

As the equations in Eq. \ref{eq:EaU1} are not differentiable, we propose differentiable approximations to these functions and introduce a trainable uncertainty calibration loss ($L_{EaUC}$) in Eq. \ref{eq:EaU2}. This loss serves as the utility-dependent penalty term within the loss-calibrated approximate inference framework for regression and continuous structured prediction tasks.
\begin{equation}\label{eq:EaU2}
  L_{EaUC} = -\log\left(\frac{n_{LC} + n_{HU}}{n_{LC} + n_{LU} + n_{HC} + n_{HU}}\right).
\end{equation}
where; \begin{align*} 
\begin{split}
    &n_{LU} = \sum_{i \in \left\{\begin{subarray}{c} ade_{i} \leq ade_{th} \\
    \quad \textrm{and} \quad c_{i} \leq c_{th}
    \end{subarray}\right\}} (1-tanh(ade_{i}))(1- c_{i}); \\
    &n_{LC} = \sum_{i \in \left\{\begin{subarray}{c} ade_{i} \leq ade_{th} \\
    \quad \textrm{and} \quad c_{i} > c_{th}
    \end{subarray}\right\}} (1-tanh(ade_{i}))(c_{i});\\
    &n_{HC} = \sum_{i \in \left\{\begin{subarray}{c} ade_{i} > ade_{th} \\
    \quad \textrm{and} \quad c_{i} > c_{th}
    \end{subarray}\right\}} tanh(ade_{i})(c_{i}); \\
    &n_{HU} = \sum_{i \in \left\{\begin{subarray}{c} ade_{i} > ade_{th} \\
    \quad \textrm{and} \quad c_{i} \leq c_{th}
    \end{subarray}\right\}} tanh(ade_{i})(1-c_{i});
\end{split}
\end{align*}

We utilize hyperbolic tangent function as bounding function to be able to scale the error and/or certainty measures to the range $[0,1]$. The intuition behind the approximate functions is that the bounded error {$tanh(ade)\rightarrow 0$} when the predictions are accurate and {$tanh(ade) \rightarrow 1$} when inaccurate. 

In our approach, the proposed EaUC loss is used as secondary loss. Eq. \ref{eq:EaU3} shows the final loss function used for continuous structured prediction task:
\begin{equation}\label{eq:EaU3}
  L_{Final} = L_{NLL} + (\beta \times L_{EaUC}).
\end{equation}
$\beta$ is a hyperparameter for relative weighting of EaUC loss with respect to primary loss (e.g. Negative Log Likelihood (NLL) in Eq. \ref{eq:EaU3}). 
\begin{algorithm}[!b]
	\caption{EaUC loss incorporated to BC model for trajectory prediction}
	\label{alg:svi_eau}
	\begin{algorithmic}[1]
	\begin{scriptsize}
		\State Given dataset $\mathcal{D}=\{X,Y\}$
		\State Given $K$ seconds of context features ($x$) and $L$ seconds of ground truth targets (y) separated by $T=0$ for each scene
		\State let parameters of MultiVariate Normal Distribution (MVN) $(\mu,\Sigma)$  \Comment{approx MVN distribution $\mathbf{Y}=\mathcal{N}(\mu, \Sigma)$ }
		\State define learning rate schedule $\alpha$
		\Repeat
		\State Sample $B$ index set of training samples; $\mathcal{D}_{m}=\left\{\left(x_i,y_i\right)\right\}_{i=1}^{B}$ \Comment{batch-size}
		\For{ $i \in B$ }
		\State $z_i=encoder(x_i)$ \Comment{encodes visual input $x_i$}
		    
		    \For{ $t \in L$} \Comment{decode a trajectory}
		    	\State $z_{i}=decoder(\hat{y}^t, z_{i})$ \Comment{unrolls the GRU}
		    	\State $\mu^t, \Sigma^t = linear(z_{i})$ \Comment{Predicts the location and scale of the MVN distribution.}
		    	\State Sample $\epsilon \sim \mathcal{N}(\mu^t, \Sigma^t)$ 
		    	\State $\hat{y_{i}}^t = \hat{y_{i}}^t + \epsilon$
		    \EndFor
		\State \label{op:unc} Calculate Average Displacement Error (ADE) \par
		\State \label{op:unc1} $ade_{i} := \frac{1}{L} \sum_{t=1}^{L} ||\hat{y_{i}^{t}}-y_{i}^{t}||_{2}.$
		\Comment ADE wrt to ground truth $y$
		
		\State \label{op:unc2} Calculate Certainty Score  \par 
		\State \label{op:unc3} $c_{i}= \log(\Pi_{t=1}^L \mathcal{N} \left(\hat{y_i}^t | \mu(y_i^{t}, x_{i}; \theta),\Sigma(y_i^{t}, x_{i}; \theta \right)))$
		\Comment{log-likelihood of a trajectory $\hat{y_i}$ in 
		\State \label{op:unc4} context $x_{i}$ to come from an expert $y_{i}$ (teacher-forcing)}
		\EndFor
		\State \label{op:calc_nac} Compute $\mathbf{n_{L C}}, \mathbf{n_{L U}}, \mathbf{n_{H C}}, \mathbf{n_{H U}}$ \Comment{ components of EaUC loss penalty}
		\State \label{op:loss} Compute loss-calibrated objective (total loss):  $\mathcal{L} = \mathcal{L}_{\textrm{NLL}} + \beta \mathcal{L}_{\textbf{EaUC}}$  
		\State Compute the gradients of loss function w.r.t to weights $w$, $\Delta\mathcal{L}_{w}$
		\State 
			Update the parameters $w$: 
			$w \leftarrow w-\alpha \Delta\mathcal{L}_{w}$ \par
		\Until{$\mathcal{L}$ has converged, or when stopped}
		\end{scriptsize}
	\end{algorithmic}\label{algorithm}
\end{algorithm}

Our approach consists of three hyperparameters, which are $ade_{th}$, $c_{th}$ and $\beta$. The details on how to set these hyperparameters are provided in Section \ref{sec:discussion}. 

Under ideal conditions, the proxy functions in Eq. \ref{eq:EaU2} are equivalent to indicator functions defined in Eq. \ref{eq:EaU1}.  This calibration loss enables the model to learn to provide well-calibrated uncertainties in addition to improved robustness.

Algorithm \ref{algorithm} shows the implementation of our method with Behavioral Cloning (BC) model \cite{BC2020}, which is the baseline model of Shifts benchmark. BC consists of the encoder and decoder stages. The encoder is a Convolutional Neural Network (CNN) which captures scene context, and decoder consists of Gated Recurrent Unit (GRU) cells applied at each time step to capture the predictive distribution for future trajectories. In our approach, we add the proposed $L_{EaUC}$ loss as secondary loss to this pipeline.

\section{Experiments and Results}
\label{sec:exp}

In our experiments, we incorporated the proposed calibration loss, $L_{EaUC}$, to two stochastic models of the recently introduced Shifts benchmark for uncertainty quantification of multimodal trajectory prediction and one Bayesian model for uncertainty quantification of regression tasks as additional evaluation of our approach. The results are provided incorporating the new loss to state-of-the art and diverse baseline architectures. 
\subsection{Multimodal Vehicle Trajectory Prediction} 

We use the real-world Shifts vehicle motion prediction dataset and benchmark~\cite{shifts2021}. In this task, distributional shift is prevalent involving real ‘in-the-wild’ distributional shifts, which brings challenges for robustness and uncertainty estimation.

Shifts dataset has data from six geographical locations, three seasons, three times of day, and four weather conditions to evaluate the quality of uncertainty under distributional shift. Currently, it is the largest publicly available vehicle motion prediction dataset containing $600,000$ scenes, which consists of both in-distribution and shifted partitions.

In Shifts benchmark, optimization is done based on NLL objective, and results are reported for two baseline architectures, which are stochastic Behavioral Cloning (BC) Model \cite{BC2020} and Deep
Imitative Model (DIM) \cite{DIM18}. We report our results incorporating the ‘\textbf{Error Aligned Uncertainty Calibration}’ loss $L_{EaUC}$ as secondary loss to Shifts pipeline as shown in Algorithm \ref{algorithm}.


Our aim is to learn predictive distributions that capture reliable uncertainty. This allows us to quantify uncertainties in the predictions during inference through sampling method and predict multiple trajectories of a target vehicle for the next $5$ \textit{sec} using data collected with $5$ \textit{Hz} sampling rate. 

During training, for each BC and DIM models, density estimator (likelihood model) is generated by teacher-forcing (i.e. from the
distribution of ground truth trajectories). The same settings of the Shifts benchmark is used: The model is trained with AdamW \cite{adamw} optimizer with a learning rate (LR) of 0.0001, using a cosine annealing LR schedule with 1 epoch warmup, and gradient clipping at 1. Training is stopped after 100 epochs in each experiment. 

During inference, Robust Imitative Planning \cite{RIP20} is applied. Sampling is applied on the likelihood model considering a predetermined number of predictions $G=10$. The top $D = 5$ predictions of the model (or multiple models when using ensembles) are selected according to their log-likelihood based certainty scores.

In this paper, we demonstrate the predictive performance of our model using the $weightedADE$ metric: \begin{equation}\label{eq:weightedADE}
\begin{footnotesize}
weightedADE_{D}(q) := \sum_{d \in D} \tilde{c}^{(d)} \cdot ADE(y^{(d)})
\end{footnotesize}
\end{equation}
In Eq. \ref{eq:weightedADE}, $\tilde{c}$ represents the confidence score for each prediction computed by applying softmax to log-likelihood scores as in Shifts benchmark.

The joint assessment of uncertainty quality and model robustness is analyzed following the same experimental setup and evaluation methodology as in Shifts benchmark, which applies error and F1 retention curves for this purpose. Lower \textit{R-AUC} (area under error retention curve) and higher \textit{F1-AUC} (area under F1 retention curve) indicate better calibration performances. In these retention curves, \textit{weightedADE} is the error and the retention fraction is based on per-prediction uncertainty score $U$. Mean averaging is applied while computing $U$, which is based on the per-plan log-likelihoods, as well as for the aggregation of ensemble results. The improvement with area under these curves can be achieved either with a better model providing lower overall error, or providing better uncertainty estimates such that more errorful predictions are rejected earlier~\cite{shifts2021}.

The secondary loss incentivizes the model to align the uncertainty with average displacement error (ADE) while training the model. Our experimental results are conducted by setting $\beta$ (see Eq. \ref{eq:EaU3}) as 200, $ade_{th}$ and $u_{th}$ (see Eq. \ref{eq:EaU2}) as 0.8 and 0.6, respectively, for both BC and DIM models. In order to scale the robustness ($ade_{i}$) and certainty ($c_{i}$) measures to a proper range for the bounding function \textit{tanh} or to be used directly, we apply post-processing. We scale the $ade_{i}$ values with $0.5$ so that we can assign samples with ADE below $1.6$ as an accurate sample. Our analysis on the Shifts dataset shows that the log-likelihood based certainty measures, $c_{i}$, of most samples are in between $[0, 100]$ range so we applied a clipping of values for the values below 0 and above 100 by setting these values to 0 and 100, respectively, and then these certainty measures are normalized to $[0, 1]$ range for direct use in Eq. \ref{eq:EaU2}. The applied post-processing steps and hyperparameter selection should be adapted according to each performed task considering initial training epochs (see Section \ref{sec:discussion} for details).

\begin{figure}[t!]
  \centering
  \includegraphics[width=0.67\linewidth,page=3]{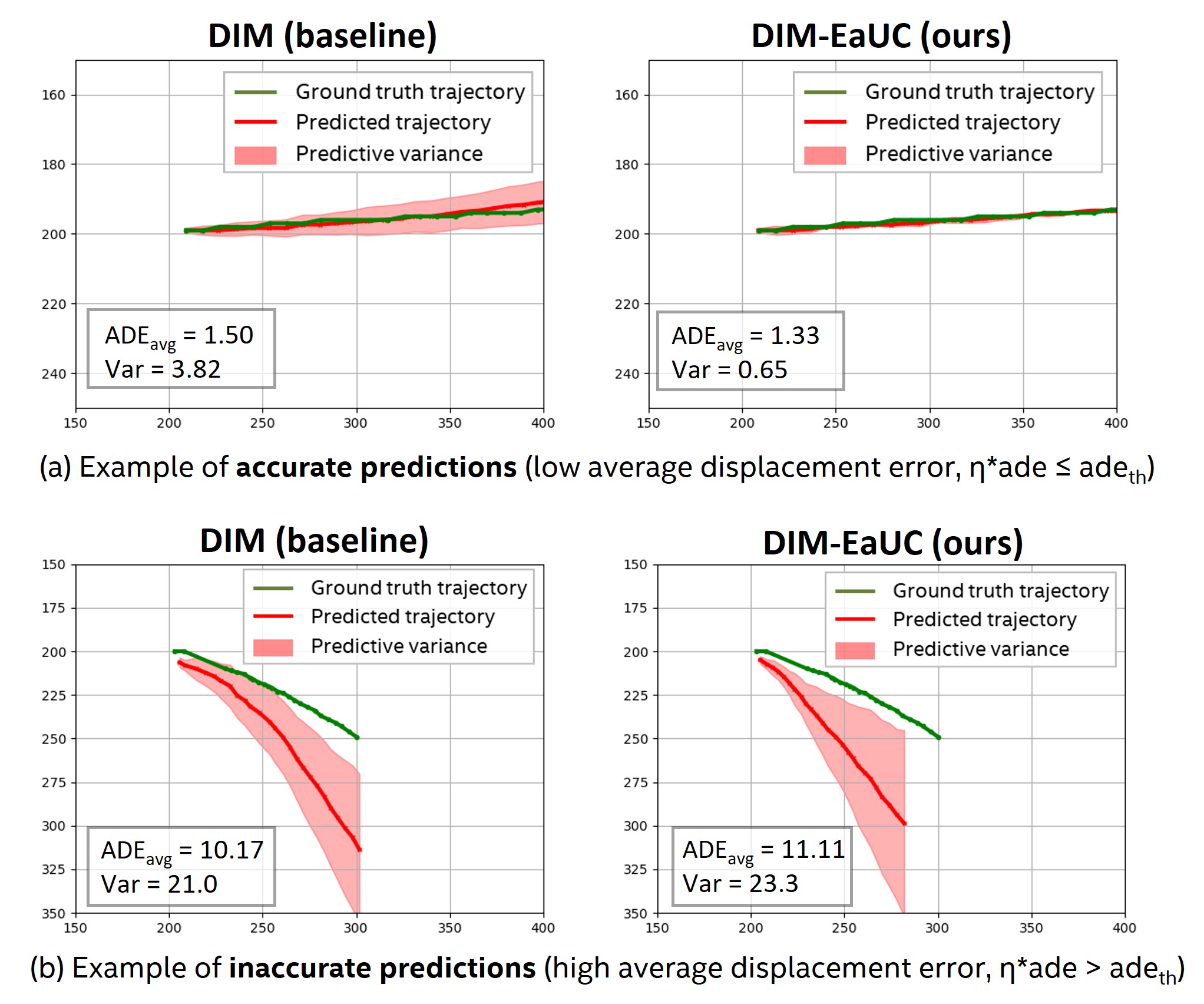}
  \caption{Figure shows correlation of accuracy with variance of the top-3 predicted trajectories based on certainty score with/without our loss for DIM baseline on Shift Dataset. With EaUC loss, we observe lower variance for accurate predictions and higher variance for inaccurate predictions. }\label{fig:variance}
\end{figure}

\textbf{Correlation analysis:} Table \ref{tab:Test10} shows the correlation between uncertainty measure and prediction error using Pearson correlation coefficient ($r$) \cite{benesty2009pearson} and the classification of samples as accurate and inaccurate using AUROC metric. \begin{itemize}
\item We observe $17.22\%$ and $19.13\%$ improvement in correlation between uncertainty and error as quantified by Pearson's $r$ when $L_{EaUC}$ loss is incorporated to BC and DIM models, respectively (Table \ref{tab:Test10}).
\item Setting accurate prediction threshold as 1.6, we observe $6.55\%$  and $8.02\%$ improvement in AUROC when $L_{EaUC}$ is incorporated to BC and DIM models, respectively, for detecting samples as accurate versus inaccurate based on the uncertainty measures (Table \ref{tab:Test10}).
\end{itemize} Fig. \ref{fig:variance} provides example cases for both accurate and inaccurate trajectory predictions for DIM baseline, which shows EaUC loss improves model calibration.
\begin{table}[!t]
	\begin{center}
		\caption{Evaluating the correlation between uncertainty estimation and prediction error with Pearson correlation coefficient $r$ and AUROC metric for binary classification of samples as accurate and inaccurate. }
		\begin{tabular}{l c c c|c c c}
		    \hline
		    \multicolumn{1}{l}{Model} & \multicolumn{3}{c}{\textbf{Pearson corr. co-eff. $\textbf{r}$} $\uparrow$} &
            \multicolumn{3}{c}{\textbf{AUROC} $\uparrow$} \\
		    \multicolumn{1}{l}{ (MA, K=1)} & In & Shifted & Full & In & Shifted & Full \\
		    \hline \hline
		    BC & 0.476 & 0.529 & 0.482 & 0.756 & 0.817 & 0.763 \\
            BC-EaUC${^*}$ & \textbf{0.557} & \textbf{0.624} & \textbf{0.565} & \textbf{0.811} & \textbf{0.833} & \textbf{0.813} \\
		    \hline
		    DIM & 0.475 & 0.522 & 0.481 & 0.754 & 0.816 & 0.761 \\
		    DIM-EaUC${^*}$ & \textbf{0.567} & \textbf{0.620} & \textbf{0.573} & \textbf{0.819} & \textbf{0.838} & \textbf{0.822} \\
		    \hline
		\end{tabular}
		\label{tab:Test10}
	\end{center}
\end{table}
\begin{table*}[!t]
    \begin{scriptsize}
	\begin{center}
		\caption{\textit{Predictive robustness (weightedADE)} and \textit{joint assessment of uncertainty and robustness performance (F1-AUC, F1@95\%, R-AUC)} of the two baselines with/without EaUC loss. Lower is better for \textit{weightedADE} and \textit{R-AUC}, and higher is better for \textit{F1-AUC, F1@95\%}.}
		\begin{tabular}{l c c c| c c c| c c c| c c c}
		    \hline
		    \multicolumn{1}{l}{Model} & \multicolumn{3}{c}{weightedADE $\downarrow$} & \multicolumn{3}{c}{R-AUC $\downarrow$} &
		    \multicolumn{3}{c}{F1-AUC $\uparrow$} &
		    \multicolumn{3}{c}{F1@95\% $\uparrow$}  \\
		    \multicolumn{1}{l}{(MA, K=1)} & In & Shifted & Full & In & Shifted & Full & In & Shifted & Full & In & Shifted & Full \\
		    \hline \hline
            BC & 1.475 & 1.523 & 1.481 & 0.423 & 0.357 & 0.416 & 0.631 & 0.646 & 0.633 & 0.848 & 0.884 & 0.852 \\
            BC-EaUC${^*}$ & \textbf{1.450} & \textbf{1.521} & \textbf{1.456} & \textbf{0.336} & \textbf{0.290} & \textbf{0.330} & \textbf{0.652}  & \textbf{0.666} & \textbf{0.654} & \textbf{0.875} & 0.884 & \textbf{0.876} \\
		    \hline
		    DIM & 1.465 & 1.523 & 1.472 & 0.418 & 0.358 & 0.411 & 0.632 & 0.647 & 0.634 & 0.851 & 0.884 & 0.855 \\
		    DIM-EaUC${^*}$ & \textbf{1.388} & \textbf{1.514} & \textbf{1.403} & \textbf{0.328} & \textbf{0.290} & \textbf{0.323} & \textbf{0.651} & \textbf{0.664} & \textbf{0.653} & \textbf{0.881} & \textbf{0.888} & \textbf{0.882} \\
		    \hline
		\end{tabular}
		\label{tab:Test4}
	\end{center}
	\end{scriptsize}
\end{table*}

\begin{figure*}[t!]
  \centering
  \includegraphics[width=0.65\linewidth] {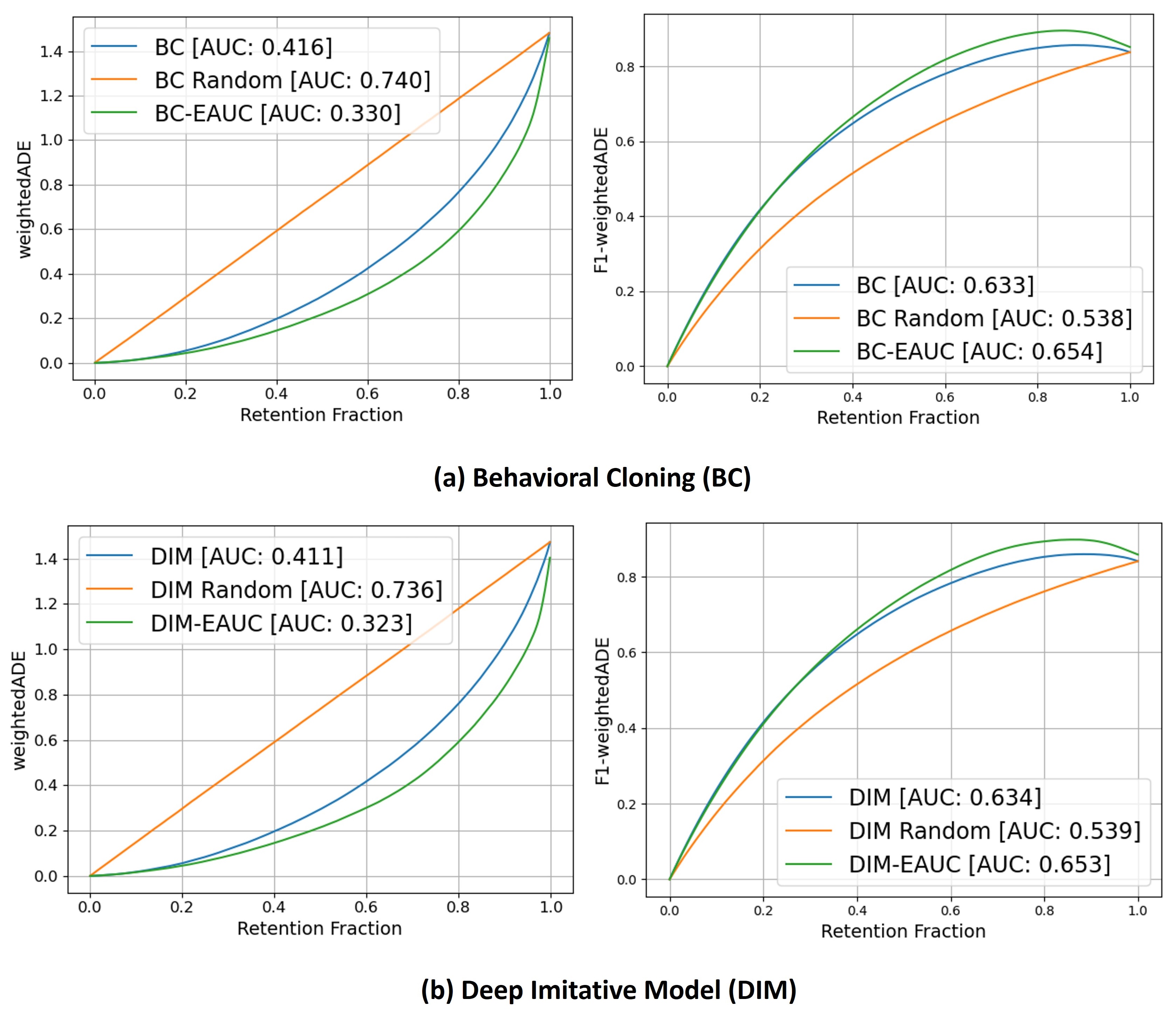}
  \caption{Error and F1-weightedADE as a function of amount of retained samples based on uncertainty scores. The retention plots are presented for robust imitative planning method applied to two baselines (BC and DIM) with/without EaUC loss using full dataset. Retention fraction is based on per-prediction request uncertainty metric $U$, which is computed with Mean Averaging (MA) of per-trajectory log-likelihood scores. Lower weightedADE and higher F1-weightedADE are better.} \label{fig:Test2}
\end{figure*}

\textbf{Quality assessment of uncertainty and model performance using retention curves:  } Table \ref{tab:Test4} shows the results for the joint quality assessment of uncertainty and robustness using \textit{R-AUC, F1-AUC, F1@95\%} metrics. Predictive performance is computed with \textit{weightedADE} metric, which is also the error metric of retention plots. Figure \ref{fig:Test2} shows error retention plots and F1-weightedADE retention plots for both BC and DIM baselines with/without our calibration loss. The results in Table \ref{tab:Test4} and Figure \ref{fig:Test2} show that: \begin{itemize}
\item There is an improvement of $1.69\%$ and $4.69\%$ for \textit{weightedADE}, and $20.67\%$ and $21.41\%$ for \textit{R-AUC} for BC an DIM models, respectively, using the full dataset. We outperform the results on two baselines providing well-calibrated uncertainties in addition to improved robustness.
\item \textit{R-AUC, F1-AUC and F1@95\%} improve for both models using all Full, In, and Shifted datasets with our $L_{EaUC}$ loss, which indicate better calibration performance using all three metrics. 
\item \textit{weightedADE} is observed to be higher for Shifted dataset compared to In dataset, which proves that error is higher for out-of-distribution data.
\end{itemize}

\textbf{Impact of weight assignment to the class of accurate\&certain samples (${LC}$) in the EaUC loss:  } In safety-critical scenarios, it is very important to be certain about our prediction when the prediction is accurate in addition to having a greater number of accurate samples.

\begin{table}[!t]
	\begin{center}
		\caption{Impact of assigning higher weights to ${LC}$ class in $L_{EaUC}$ loss on robustness and calibration performance. }
		\begin{tabular}{l c|c}
		    \hline
		    \multicolumn{1}{l}{Model (MA, K=1)} & \multicolumn{1}{c}{weightedADE $\downarrow$} &
            \multicolumn{1}{c}{R-AUC $\downarrow$} \\
		    \hline \hline
		    BC-EaUC & 1.880 & 0.355 \\
            BC-EaUC${^*}$ & \textbf{1.456} & \textbf{0.330} \\
		    \hline
		    DIM-EaUC & 1.697 & 0.344 \\
		    DIM-EaUC${^*}$ & \textbf{1.403} & \textbf{0.323} \\
		    \hline
		\end{tabular}
		\label{tab:Test8}
	\end{center}
\end{table}
 
To incentivize the model during training more towards accurate predictions and be certain, we experimented by giving higher weightage to the class of $LC$ samples while computing Eq. \ref{eq:EaU2}. Eq. \ref{eq:EaU5} shows how we assign high weights to these samples in our loss where $\gamma > 1$. We force the algorithm to better learn the samples of this class and obtain both improved calibration (well-calibrated) and robustness (lower weightedADE) with this approach as shown in Table \ref{tab:Test8}.

\begin{equation}\label{eq:EaU5}
L_{EaUC} = -\log\left(\frac{(\gamma \cdot n_{LC}) + n_{HU}}{(\gamma \cdot n_{LC}) + n_{LU} + n_{HC} + n_{HU}}\right).
\end{equation}

BC-EaUC/DIM-EaUC and BC-EaUC${^*}$/DIM-EaUC${^*}$ denote the results according to Eq. \ref{eq:EaU2} and Eq. \ref{eq:EaU5}, respectively. BC-EaUC${^*}$ and DIM-EaUC${^*}$ provide better performance in terms of robustness (\textit{weightedADE}) and model calibration (\textit{R-AUC}) compared to BC-EaUC and DIM-EaUC, hence all the experiments in this section are reported applying our loss according to Eq. \ref{eq:EaU5}, where $\gamma=3$, for both models. Our analysis showed that increasing $\gamma$ value above some threshold causes degradation in performance. Table \ref{tab:Test8} shows the results on Full dataset.

Another observation in this experiment is that even though BC-EaUC and DIM-EaUC provide slightly higher \textit{weightedADE} compared to their corresponding baselines (BC and DIM in Table \ref{tab:Test4}), they provide better quality uncertainty estimates as quantified by \textit{R-AUC}. 


\textbf{Study with ensembles:  } Table \ref{tab:Test5} shows that using ensembles, better performances are achieved for both predictive model performance and calibration performances with the joint assessment of uncertainty and robustness using both BC and BC-EaUC${^*}$. EaUC together with Ensembles provides the best results.
\begin{table*}[!t]
    \begin{scriptsize}
	\begin{center}
		\caption{Impact analysis of $L_{EaUC}$ loss on ensembles using the aforementioned metrics for robustness and joint assessment}
		\begin{tabular}{l c c c| c c c| c c c| c c c}
		    \hline
		    \multicolumn{1}{l}{Model} & \multicolumn{3}{c}{weightedADE $\downarrow$} & \multicolumn{3}{c}{R-AUC $\downarrow$} &
		    \multicolumn{3}{c}{F1-AUC $\uparrow$} &
		    \multicolumn{3}{c}{F1@95\% $\uparrow$}  \\
		    \multicolumn{1}{l}{(MA)} & In & Shifted & Full & In & Shifted & Full & In & Shifted & Full & In & Shifted & Full \\
		    \hline \hline
            BC (K=1) & 1.475 & 1.523 & 1.481 & 0.423 & 0.357 & 0.416 & 0.631 & 0.646 & 0.633 & 0.848 & 0.884 & 0.852 \\
            BC (K=3) & \textbf{1.420} & \textbf{1.433} & \textbf{1.421} & \textbf{0.392} & \textbf{0.323} & \textbf{0.384} & \textbf{0.639} & \textbf{0.651} & \textbf{0.640} & \textbf{0.853} & \textbf{0.887} & \textbf{0.857}  \\
		    \hline
            BC-EaUC${^*}$ (K=1) & 1.450 & 1.521 & 1.456 & 0.336 & 0.290 & 0.330 & 0.652 & 0.666 & 0.654 & 0.875 & 0.884 & 0.876 \\
            BC-EaUC${^*}$ (K=3) & \textbf{1.290} & \textbf{1.373} & \textbf{1.300} & \textbf{0.312} & \textbf{0.271} & \textbf{0.308} & \textbf{0.647} & \textbf{0.659} & \textbf{0.648} & \textbf{0.890} & \textbf{0.898} & \textbf{0.891} \\
		    \hline 
		\end{tabular}
		\label{tab:Test5}
	\end{center}
	\end{scriptsize}
\end{table*}

\subsection{UCI Regression} 
\label{regression}
We evaluate our method on UCI~\cite{Dua2019} datasets following \cite{immer2021scalable} and \cite{gal2016dropout}. We use the Bayesian neural network (BNN) with Monte Carlo dropout~\cite{gal2016dropout} approximate Bayesian inference. In this setup, we use neural network with two hidden layers fully-connected with 100 neurons and a ReLU activation. A dropout layer with probability of $0.5$ is used after each hidden layer, with $20$ Monte Carlo samples for approximate Bayesian inference. Following \cite{antoran2020depth}, we find the optimal hyperparameters for each dataset using Bayesian optimization with HyperBand~\cite{falkner2018bohb} and train the models with SGD optimizer and batch size of $128$. The predictive variance from Monte Carlo forward passes is used as the uncertainty measure in the EaUC loss function. BNN trained with a secondary EaUC loss yields lower predictive negative log-likelihood and lower RMSE on multiple UCI datasets as shown in Table 5. Fig. 4 shows the model error retention plot based on the model uncertainty estimates for UCI boston housing regression dataset. BNN trained with loss-calibrated approximate inference framework using EaUC loss yields lower model error as most certain samples are retained with $15.5\%$ lower AUC compared to baseline BNN and $31.5\%$ lower AUC compared to Random retention (without considering uncertainty estimation). These results demonstrate that EaUC enables to obtain informative and well-calibrated uncertainty estimates from neural networks for reliable decision making.

\begin{minipage}{1\textwidth}
    \begin{minipage}[b]{0.48\textwidth}
    \begin{footnotesize}
    \text{\textbf{Table 5}  }{Avg. predictive negative log-likelihood (NLL)(lower is better) and test root mean squared error (RMSE)(lower is better) on the UCI regression benchmark.}
    \begin{tabular}{lccccc}
    \hline
    \multirow{2}{*}{UCI } & \multicolumn{2}{c}{Test NLL $\downarrow$}                                 & \multirow{10}{*}{} & \multicolumn{2}{c}{Test RMSE $\downarrow$}                                                 \\ \cline{2-3} \cline{5-6} 
                                   Dataset   & \textbf{BNN}    & \textbf{\begin{tabular}[c]{@{}c@{}}BNN \\ (EaUC)\end{tabular}} &                    & \textbf{BNN} & \textbf{\begin{tabular}[c]{@{}c@{}}BNN \\ (EaUC)\end{tabular}} \\ 
    \hline
    \textbf{boston}                   & 2.424           & \textbf{2.203}                                                 &                    & 2.694        & \textbf{2.317}                                                 \\ 
    \textbf{concrete}                 & 3.085           & \textbf{3.016}                                                 &                    & 5.310        & \textbf{4.580}                                                 \\ 
    \textbf{energy}                   & 1.675           & \textbf{1.643}                                                 &                    & 0.973        & \textbf{0.671}                                                 \\ 
    \textbf{yacht}                    & 2.084           & \textbf{1.684}                                                 &                    & 2.024        & \textbf{0.787}                                                 \\  
    \textbf{power}                    & 2.885           & \textbf{2.829}                                                 &                    & 4.355        & \textbf{4.018}                                                 \\ 
    \textbf{naval}                    & \textbf{-4.729} & -4.294                                                         &                    & 0.001        & 0.001                                                          \\ 
    \textbf{kin8nm}                   & -0.930          & \textbf{-0.967}                                                &                    & 0.086        & \textbf{0.074}                                                 \\  
    \textbf{protein}                  & 2.888           & \textbf{2.851}                                                              &                    & 4.351        & \textbf{4.181}                                                              \\ \hline
    \end{tabular}
    \label{tab:uci}
    \end{footnotesize}
    \end{minipage}
\hskip 20pt 
    \begin{minipage}[b]{0.4\textwidth}
       \includegraphics[width=1\linewidth]{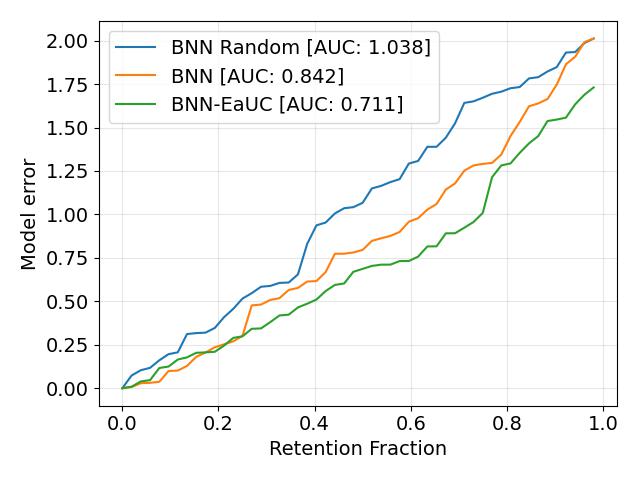}
       \begin{footnotesize}
       \text{\textbf{Fig. 4    }}{ Error retention plot for BNN on UCI boston housing dataset to evaluate the quality of model uncertainty (lower AUC is better). }\label{fig:uci_retention}
       \end{footnotesize}
    \end{minipage}
\end{minipage}

\subsection{Hyperparameter selection}
\label{sec:discussion}
Our solution requires three hyperparameters, which are the thresholds $ade_{th}$ and $c_{th}$ to assign the sample to appropriate category of prediction (certain \& accurate; certain \& inaccurate; uncertain \& inaccurate; uncertain \& accurate) as well as the $\beta$ parameter to assign relative weighting to our secondary loss compared to primary loss. Initially we train the model without secondary calibration loss for few epochs, then we analyze both ADE and log-likelihood values to set the thresholds required for calibration loss following the same strategy as \cite{AvUC20}. We perform grid search in order to tune the threshold parameters. Post-processing is applied on certainty measures to set the values between 0 and 1 range for direct use of these measures in our loss function. $\beta$ is chosen such that the introduced secondary loss contributes significantly to the final loss, we select this value such that the secondary loss is at least half of the primary loss. For the regression experiments, we found the optimal hyperparameters for each dataset using Bayesian optimization with HyperBand \cite{falkner2018bohb}.

\section{Conclusions}
Reliable uncertainty estimation is crucial for safety-critical systems towards safe decision making. In this paper, we proposed a novel error aligned uncertainty (EaU) optimization method and introduced differentiable EaU calibration loss that can be used as a secondary penalty loss to incentivize the models to yield well-calibrated uncertainties in addition to improved robustness. 
To the best of our knowledge, this is the first work to introduce a trainable uncertainty calibration loss function to obtain reliable uncertainty estimates for time-series based multi-variate regression tasks such as trajectory prediction. We evaluated our method on real-world vehicle motion prediction task with large-scale Shifts dataset involving distributional shifts. We also showed that the proposed method can be applied to any regression task by evaluating on UCI regression benchmark. The experimental results shows the proposed method can achieve well-calibrated models yielding reliable uncertainty quantification in addition to improved robustness even under real-world distributional shifts. The EaU loss can also be utilized for post-hoc calibration of pretrained models, which we will explore in future work. We hope our method can be used along with well-established robust baselines to advance the state-of-the-art in uncertainty estimation benefiting various safety-critical real-world AI applications.

\subsubsection*{Acknowledgements} 
This project has received funding from the European Union’s Horizon 2020 research and innovation programme under grant agreement No 956123. This research has also received funding from the Federal Ministry of Transport and Digital Infrastructure of Germany in the project Providentia++ (01MM19008F).




\clearpage
%
%
\bibliographystyle{splncs04}
\bibliography{egbib}
\end{document}